\documentclass[letterpaper, 10 pt, conference]{ieeeconf}  

\IEEEoverridecommandlockouts                              

\usepackage{graphicx} 

\title{\LARGE \bf
Social Robots for People with Developmental Disabilities: A User Study on Design Features of a Graphical User Interface}

\author{Xiaodong Wu and Lyn Bartram
\thanks{This work was supported by JDQ Inc. and Mitacs through the Mitacs Accelerate program.}
\thanks{Both authors are with the School of Interactive Arts and Technology (SIAT), Simon Fraser University, Surrey, BC V3T 0A3, Canada
      {\tt\small xavierw@sfu.ca} and {\tt\small lyn@sfu.ca}}%
}

\begin{document}

\maketitle
\thispagestyle{empty}
\pagestyle{empty}

\begin{abstract}

Social robots, also known as service or assistant robots, have been developed to improve the quality of human life in recent years. The design of socially capable and intelligent robots can vary, depending on the target user groups. In this work, we assess the effect of social robots' roles, functions, and communication approaches in the context of a social agent providing service or entertainment to users with developmental disabilities. In this paper, we describe an exploratory study of interface design for a social robot that assists people suffering from developmental disabilities. We developed series of prototypes and tested one in a user study that included three residents with various function levels. This entire study had been recorded for the following qualitative data analysis. Results show that each design factor played a different role in delivering information and in increasing engagement. We also note that some of the fundamental design principles that would work for ordinary users did not apply to our target user group. We conclude that social robots could benefit our target users, and acknowledge that these robots were not suitable for certain scenarios based on the feedback from our users.

\end{abstract}

\section{INTRODUCTION}

Developmental disabilities are a set of permanent and severe problems that are challenging many people nowadays. As the \textit{2012 Canadian Survey on Disability (CSD)} has revealed,  ``160,500 (0.6\% of Canadian adults) were identified as having a developmental disability." This survey also indicated that 90\% of adults with a developmental disability needed assistance with some kind of everyday activity, and 72.7\% of them reported some degree of unmet need for at least one of these activities \cite{Bizier2015}. Therefore, our study could be significant for designing a supportive social robot that can improve the life qualities of people having daily challenges because of their developmental disabilities.  

People with developmental disabilities often require more assistance to learn, understand or express information than others, because developmental disabilities can dreadfully affect their language and social skills \cite{DSOntario2016}. We believe that social robots, often considered as assistive robots, can improve the life quality of individuals with developmental disabilities. Since these people talk and behave differently from ordinary people, and many of them have cognitive difficulties, social robots that are designed to communicate with them must have a robust interaction model to achieve a high communication quality. A graphical user interface (GUI) is an effective way to present information. Our study followed a previous investigation on how caregivers communicate with their residents with developmental disabilities. Through this process, we analyzed the interaction patterns of both sides. The results obtained from this pretest study helped us with our research and design which we are presenting in this article. We gained an insight into our users' needs, and cooperated with caregivers to find practical solutions to the challenges that both users and caregivers are facing every day.

The application of social robots to elderly care was discussed in Broekens et al.'s article \cite{Broekens2009}. This article introduces a significant study as there is an increasing necessity for technologies that can improve the life quality of the elderly. Social robots are useful in eldercare, because they can be a gateway to elderly interfacing with digital technology, and they can also offer company to the elderly. Nevertheless, the effectiveness of this application has not been investigated much in the past. Broekens et al. compared several different social robots to investigate the effects of assistive social robots on the health of the elderly. The results of this comprehensive research indicate that assistive social robots have many functions and effects, including increasing health by decreasing level of stress, decreasing loneliness, increasing communication activity with others and improving the sense of happiness.

This paper presents the exploratory findings we had obtained after testing a GUI prototype that we designed and implemented on \textit{Aether}, an intelligent sociable service robot developed by JDQ Inc. In the past, our pilot survey suggested that most of the elderly reported a liking for robots. This research shows the validity of using social robots for people with developmental disabilities. The user study provided new perspectives for us to scrutinize users' affect and emotion that the robot had triggered. We conclude that the implementation of the GUI played a crucial role in the organic unity of the robot system.

\section{RELATED WORK}

Previous studies on social robotics have confirmed users' acceptance of therapeutic social robots in everyday scenarios, acknowledging the fact that users' satisfaction is determined by the user-technology fit \cite{Baisch2017}.

When it comes to robot design, there are always moments that we have to decide whether or not to make an aspect more humanoid. Walters et al. \cite{Walters2008} suggest that users tend to be more intrigued by robots with characteristics and behaviours that differ from humans. However, this theory, in the field of social psychology, may not apply to the very user group which we are focusing on. Even for one single user group, there can be noticeable individual differences in terms of personalities and temperaments. Thus, decisions to attach personality traits to social robots have to be made cautiously with a subsequent assessment conducted. 

An extant study \cite{Mohd2012} points out that users are more likely to struggle with understanding GUI if the cognitive psychological principles and theories are not applied to the design properly or sufficiently, by conducting a series of empirical experiments based on theories such as Schema Theory and Gestalt Law. These experiments consist of multiple GUI design aspects including proximity, colour, placement of content, and type of icons (familiar vs. unfamiliar).

As part of our robot GUI design, we first need to determine the manifestation of the virtual agent, if any. Since our robot has a physical presence already, instead of only being a pure personified character in a computer screen, we wanted to augment the embodiment of the graphic imagery in a physical semi-humanoid robot. Wang et al. \cite{Wang2017} argue that 3D face stimuli expedite children's recognition of facial expressions, based on an experiment on 71 children aged between 3 and 6. This study had been inspiring to our project, partly because of the closeness of IQ between our user groups. However, our tests of different portrayals of a given character did not show any evident disparity regarding users' acceptance, whether it was 2D or 3D.

As the relationship between users' engagement and the social attributes of assistive robots is the primary topic of this study, we need to create a robust performance benchmark to assess the impact of each attribute. In the area of HRI, there are various benchmarks to evaluate the performance of a robot system. Some of them mainly focus on such technical aspects as scalability and imitation \cite{Feil-Seifer2007}. To us concentrating on the sociability of assistive robots, we determined to construct an efficient and practical benchmark from a less technical perspective. We aimed to employ this evaluation model to detect and measure \textit{Aether}'s impact on both individuals with developmental disabilities and their caregivers. The following fundamental usability criteria had been taken into our evaluation of the overall effectiveness of our design: error prevention, efficiency, effectiveness, and user satisfaction. The interesting challenge was the tension between the expectations and the allowances of the effect the robot has on the caregiver practices.

\section{METHODOLOGY}

 This study was divided into three separate stages based on the types of interaction: human-human interaction, human-computer interaction (HCI), and human-robot interaction (HRI). At least two cameras, including a handheld camera and a webcam, were used to capture the entire study process alongside participants' facial emotions and gestural responses.

\subsection{Robot Design Considerations}

\begin{figure}[hbt]
\centering
\includegraphics[width=2.2in]{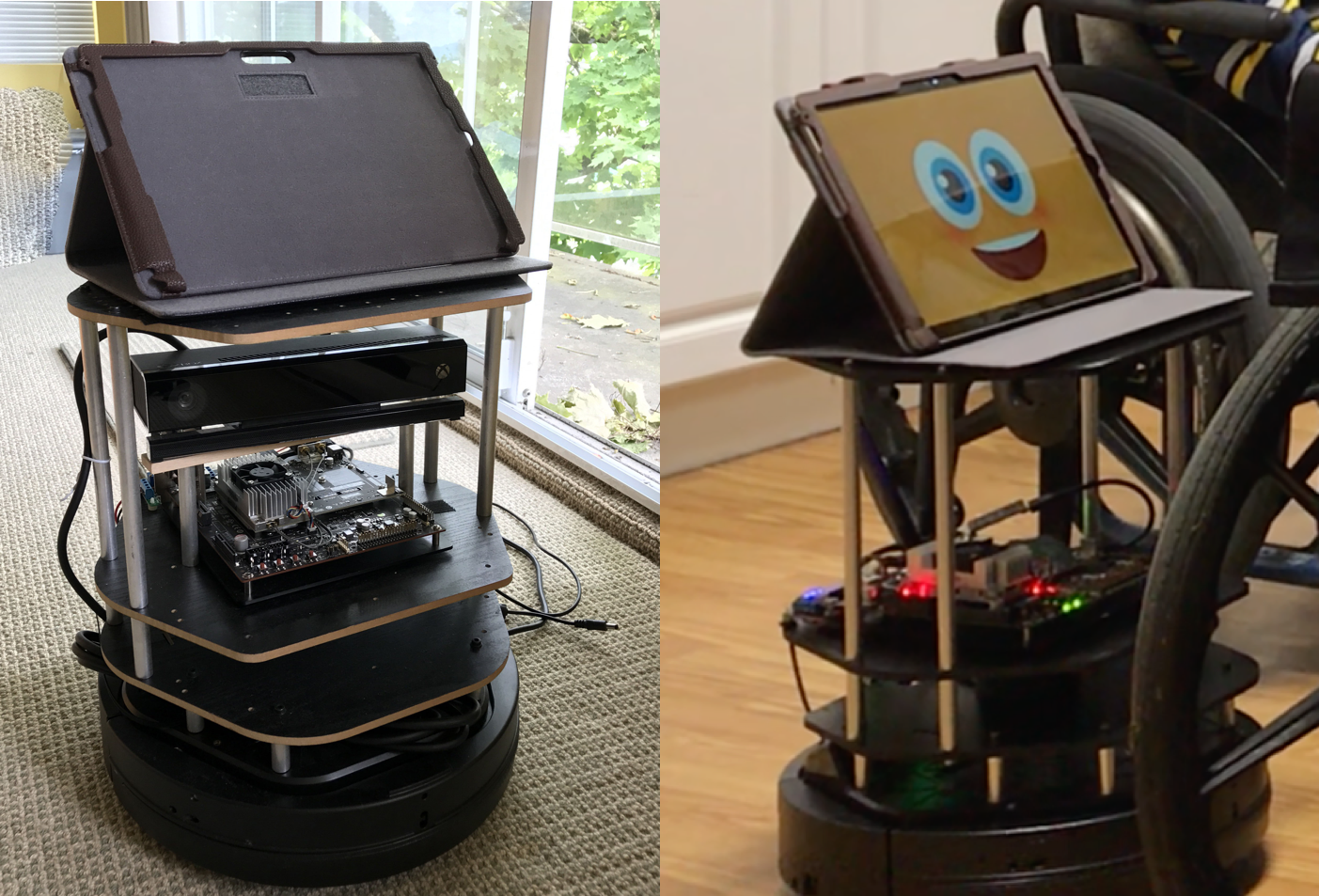}
\caption{The setup of \textit{Aether} without (left) and with (right) the display}
\label{fig:robotsetup}
\end{figure}

As illustrated in Fig. \ref{fig:robotsetup}, a monitor was installed on Kobuki \cite{Kobuki2015}, the mobile robot base, along with other sensory and computing components. A small yet powerful computing unit was installed on the base, running ROS on Ubuntu 17.04. \textit{Aether} could thus be controlled remotely via the SSH protocol. The entire system was about 20 inches tall, with a replaceable monitor mount to ensure that we could change the display module based on our design needs in the future.

\subsubsection{Personification of Robot}

Many researchers have made endeavours to categorize robot personality, and there was evidence that people spend more time gazing at robots which behave actively, suggesting that users tend to be more interested in extrovert robots \cite{Mubin2014}. This coheres with the psychological discovery \cite{Hendrick1971} that the introvert has more heterogeneous relationships with people: the extrovert is likely to be only attracted to gregarious persons because of their similarities, whereas there is no negative indicator when the introvert interact with the extrovert. However, it is also worth pointing out that previous studies on personality psychology have come to two outwardly paradoxical theories: similarity attraction and complementary attraction \cite{Brandon2012}. The similarity attraction theory implies that individuals tend to get attracted by a homogeneous group of similar personalities, whereas the complementary attraction theory suggests the opposite.

Whether it is similarity attraction and complementary attraction, we had to explore and revalidate this by ourselves for the following reasons. Firstly, our users might behave differently from their true internal personalities due to cognitive impairments. We could not conclude a correlation between their personalities and their engagement with robot. Secondly, we had created a catalogue of existing residents' profiles, including their habits, personal preferences, always-up-to-date schedules, etc based on previous observations and interviews. Utilizing these data could improve the quality of human-robot interaction (HRI) as the robot would be able to pinpoint users' needs and desires. Lastly, we had not determined whether this robot can serve only one or multiple users at the same time. As a virtual agent, the robot was being considered to be capable of interacting with all types of user. Therefore we needed to set up a default personality ``value" for the robot in the event that a new user's profile has not been indexed in the database. Therefore, we would not make the robot fully personalized until the user group gets larger at this stage.

\subsubsection{Robot GUI States}

``Robot GUI state" refers to the state variable that shows the current status of a robot in a humanoid way. It lays the foundation for effective interaction and high engagement. Symbols and derived animations were used to reflect these different states. The most common states include: ``Smiling", ``Thumbs-up", ``Cheerful", ``Sleeping", and ``Neutral." One same state could be repeatedly used for different scenarios. For example, the ``cheerful" state can be triggered when someone's birthday comes, or simply when the caregiver is going to give the resident a ride outside. It is vital to integrate emotion expressions to the robot's personification, because of the significantly positive correlation between robot's capability of expressing emotions and the quality of human-robot interactive communication \cite{Leyzberg2011}. There are three phrases to test these GUI states.

\begin{itemize}
  \item {\textbf{Evaluate the frequency of each state based on the most common scenarios identified previously}}

  Previously, we had identified the most common and essential scenarios that clients underwent every day. These scenarios included medication reminders, activity reminders, daily tasks, etc. There are many other states besides these common ones listed above. Not all states need to be implemented due to users' cognitive challenges. They can only memorize and learn to identify a few states, so diminishing users' learning curve is significant.
  
  \item {\textbf{Test the effectiveness of the graphics and animations}}
 
Keeping a good consistency in graphical styles is the prerequisite of achieving an accurate evaluation of different states. When a user fails to understand a symbol, it is likely that the very image/video being used is not clear enough, then we will need to find alternative resources to retest that state.

  \item {\textbf{Assess the necessity of using animations for each state}}

Some of the states can be more effectively represented using static images.

\end{itemize}

\subsubsection{Symbol}

One significant goal of our design was to reduce users' anxiety. This was also an essential prerequisite for keeping users engage in the interaction with the robot. Graphic symbol as one of the key factors to the visual design substantially affected users' perception of information that the robot is delivering. As textual information was not be used in our GUI design due to the verbal challenges of our user group, symbols and icons indicating tasks and objects were undoubtedly the primary approach to visually transmitting information.

Nevertheless, symbols might not always be a positive factor. Some users were fully capable of understanding verbal directives, and thus reading graphics in addition to verbal prompts is unnecessary. It might cause more confusion or anxiety because of this extra means of instruction. Also, although most of the users could understand the symbols with which they have been trained previously, they might dislike these icons. The direct outcome of using graphics to them would be that they would not even look at the graphics. Therefore, it was imperative to classify users into two basic groups based on their perceptive capabilities and interest in graphics before applying symbols.

\subsubsection{Layout}

We held an assumption that the dominant area on the display should be reserved for symbols and icons instead of the robot's virtual face, and this guess was confirmed by a caregiver. We held an assumption that the residents tended to focus on things that are simple and large, and like most of the people they read from top to bottom, left to right. 

\subsubsection{Animation}
Similar to symbols, there was also a noticeable individual difference in terms of attention to animations. We were interested to find out whether using animation would bring positive effect to users' understanding and acceptance of content.

\subsubsection{Colour}
It had been affirmed that most of the residents have an awfully hard time reading white symbols on a black background. Previous research from psychiatrists and visual analytical scholars had also found that different colour palettes produce effects on users' perceptions and trigger emotions differently \cite{Song:2017fj}.

\begin{figure}[hbt]
\centering
\includegraphics[width=3.4in]{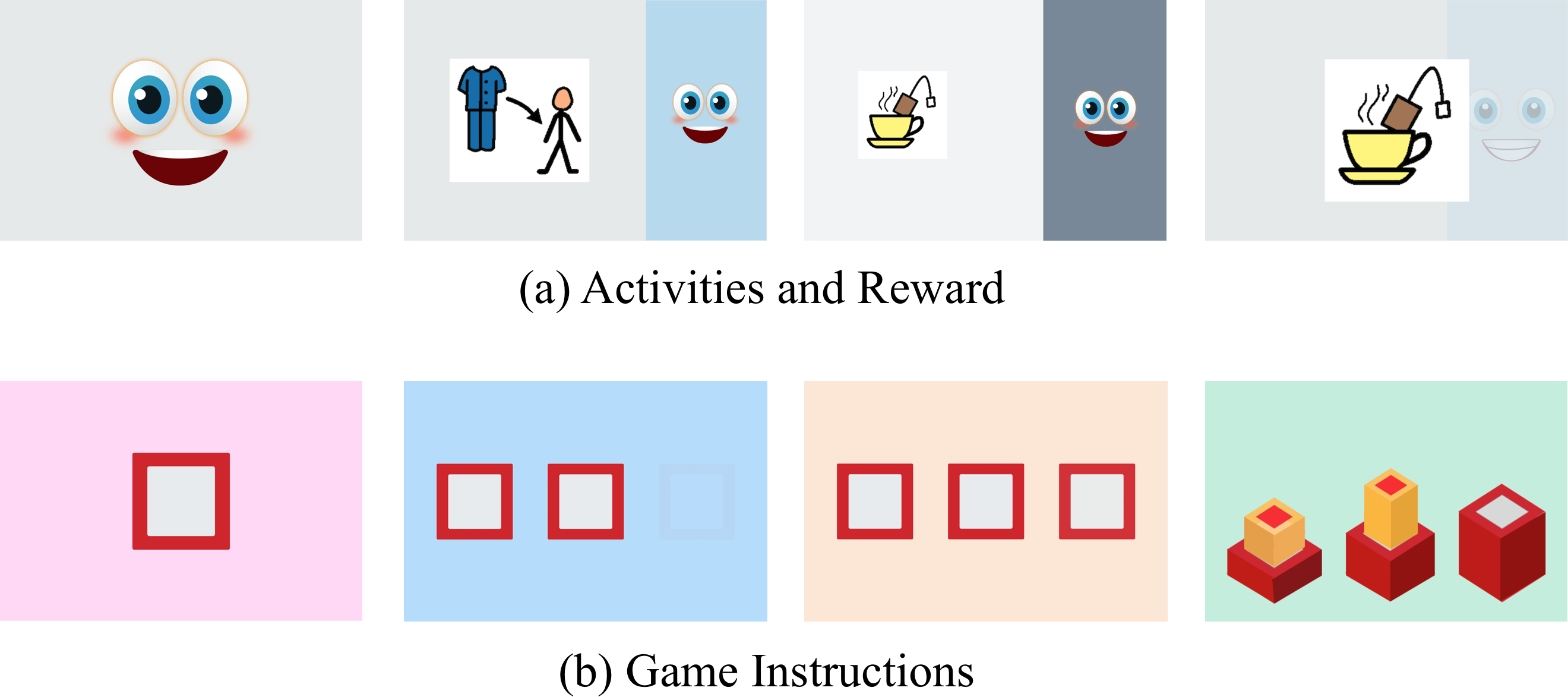}
 \caption{Sample illustrations from the prototype}
\label{fig:prototype1}
\end{figure}

\subsection{Participants}

The group of participants in this study consisted of three residents (referred as "clients" or "users" in this article)  at a local group home managed by Developmental Disabilities Association (DDA). All participants were female adults with various degrees of developmental disabilities, which meant they also had different levels of cognitive abilities, and verbal and physical skills. Rather than generalizing our assumptions and conclusions. we mainly aimed to find the significant factors differentiating between different types of interaction due to the small sample size.

Resident 1 had rather limited verbal and physical skills. She was always in a wheelchair with desk arm. Resident 1 also had some degree of delusion with symptoms such as giggling abruptly by herself. She had receptive language difficulties resulting in her restricted comprehension of simple words and phrases like ``TV".

Resident 2 had good verbal skills, but she was not very responsive all the time. There were occasions when she could answer some simple questions with full sentences, but only with caregivers whom she was familiar with. Resident 2 had several typical symptoms of dementia, including problems with focusing and paying attention, immature judgment and decision making, and limited reasoning skills. She could stare at an object for a long period of time, but it was difficult for researchers to understand if she was indeed focusing on the object or just in a trance.

Resident 3 was able to verbally communicate with both companions and strangers. She could be a bit over-talkative occasionally and tended to give answers that can please questioners. Although she had to sit in her wheelchair all the time, she was still able to provide information using signs, gestures and facial expressions. Resident 3 did not have good executive functions, or planning and sequencing skills. She also had challenges in short-term memory, which was often a sign of dementia. Due to the declined thinking and reasoning skills, oftentimes she had confusion and troubles in solving problems and following storylines.

\subsection{Usability Test}
It would be a challenge if we were going to inquire feedback of residents, because of not only the verbal skills but also their cognitive abilities. For some of them, they were just going to choose whatever comes the first as ``the best one" when you present a few selections. Therefore, we would move our focus to clients with higher cognitive and verbal abilities, and also caregivers. All experiments and interviews were conducted on an individual basis. The focus groups approach were not be adopted to avoid groupthink or errors caused by users' incapability to convey their real feelings.

\subsection{User Study Procedure}

The first part of the study was completely observational. A caregiver was asked to talk to each of the residents for about 3 minutes. There was not any specific topic or question that the caregiver had to mention. In contrast, a caregiver could bring up any topic that she/he might feel appropriate to attract residents' interest. The conversation, along with the study environment, was kept the same as the participants' daily experience. Therefore, there could be distractions or unexpected events occurring, and the researcher would not intervene in the event that anything happens. The researcher hid behind the participant to record the study.

\begin{figure}[hbt]
\centering
\includegraphics[width=3.42in]{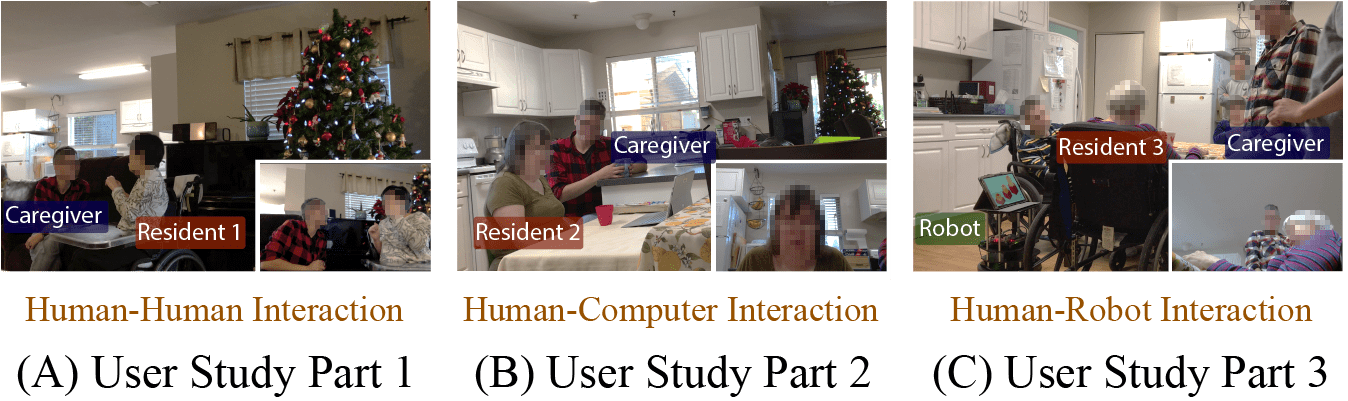}
\caption{Three stages of the user study, corresponding to three types of interaction}
\label{fig:userstudy}
\end{figure}

In the second part, a 12-in Microsoft Surface 5 computer (11.50 x 7.9 inches) was placed on a table in front of the participant with the keyboard detached. A three-minute GUI prototype started to play automatically shortly after the researcher initiated the demonstration. The GUI prototype contained a series of tasks and rewards as an imitation of the ``To-do Board" being used at DDA group homes. A caregiver sat next to the participant but remained silent the entire time to help reduce participant's  anxiety. The caregiver may give a gestural prompt to help the resident focus on the display if she gets distracted.

In the third part, the computer was mounted on the robot, playing the same GUI demo as before. We then conducted a series of Wizard of Oz (WOZ) experiments. The WOZ technique is a prevailing solution to iterative prototyping \cite{McTear2016}. It is commonly used to test the prototype to discover the drawbacks of the current design, by observing users' reactions and acceptance. The researcher hid behind the participant while controlling the movement of the robot. One primary goal of this section was to explore the effect of the robot's limited physical abilities by observing user's emotional changes, especially when the robot was approaching to and leaving the user. To further assess the impacts of design elements on HRI, we developed a specific GUI prototype to provide board game instructions to one of our participants who had the highest verbal skills and cognitive abilities amongst all. We controlled the robot to approach the user and stay about 10 inches from her. A \textit{Melissa \& Doug} shape sequence sorting set was jumbled and prepared on the table in front of the user. This simple educational toy was initially designed for children of ages 3- to 7-years old to improve their reasoning and math skills. As this GUI demo showed step-by-step instructions to the game, it would be possible to evaluate the effectiveness of HRI based on objective metrics such as completion results. We could also investigate which part of the design has flaws according to the ``roadblocks" that the user encounters.

\subsection{Follow-up Survey}

As it was difficult to create a single benchmark to numerically assess the design factors that we are investigating, we asked two caregivers to fill out a questionnaire consisting of eight concise questions. We also conducted an open-ended interview with these two caregivers to get more detailed and specific perspectives and suggestions for the user study. Both caregivers were female full-time employees with years of experience of caring for residents with developmental disabilities, and they observed the entire user study.

The eight questions are as follows: On a scale of 1 - 10 with 10 being the most effective or positive, how would you --
\begin{itemize}
\item Q1: describe the overall experience/process of delivering instructions?
\item Q2: describe the verbal instructions?
\item Q3: describe the graphic illustrations (shapes, colours, etc)?
\item Q4: describe the animations and transitions between scenes?
\item Q5: evaluate users' interest and engagement in this demo?
\item Q6: evaluate the social attributes of this robot (during this demonstration)? 
\item Q7: assess the friendliness of this robot?
\item Q8: assess the reality of this robot? (Does the appearance look real? Do these shapes and graphics correspond the reality?)
\end{itemize}

\subsection{Data Analysis}

We chose qualitative, instead of quantitative, data analysis as our primary investigation technique for two reasons. First, the small sample size of our user group had made it impractical to draw accurate generalized conclusions. Due to the nature of this study, we are unlikely to have a much larger user group in the future. Second, our results for the user study had multi-dimensionally crossing relations that would lead to more meaningful clues using qualitative coding rather than descriptive analysis.

All video and audio recordings had been compiled into a single video file, showing each stage of the study from two different perspectives on one screen. This compiled data file was then imported into NVivo, a qualitative data analysis (QDA) software package \cite{nvivo}, to process and code textual and multimedia data. Each key frame of the video had been noted with a concise description. Then, along with user's response being classified, each of these highlighted frames was categorized based on the form of communication, the type of information, and the kind of interaction. By running several matrix coding queries in NVivo, a comprehensive node matrix was created to cross-tabulate the observations of this study at each stage. The results of this node matrix are presented in Table \ref{tbl:results2}.

\section{Results}


\subsection{Survey \& Interview}

\begin{figure}[hbt]
\centering
\includegraphics[width=3.4in]{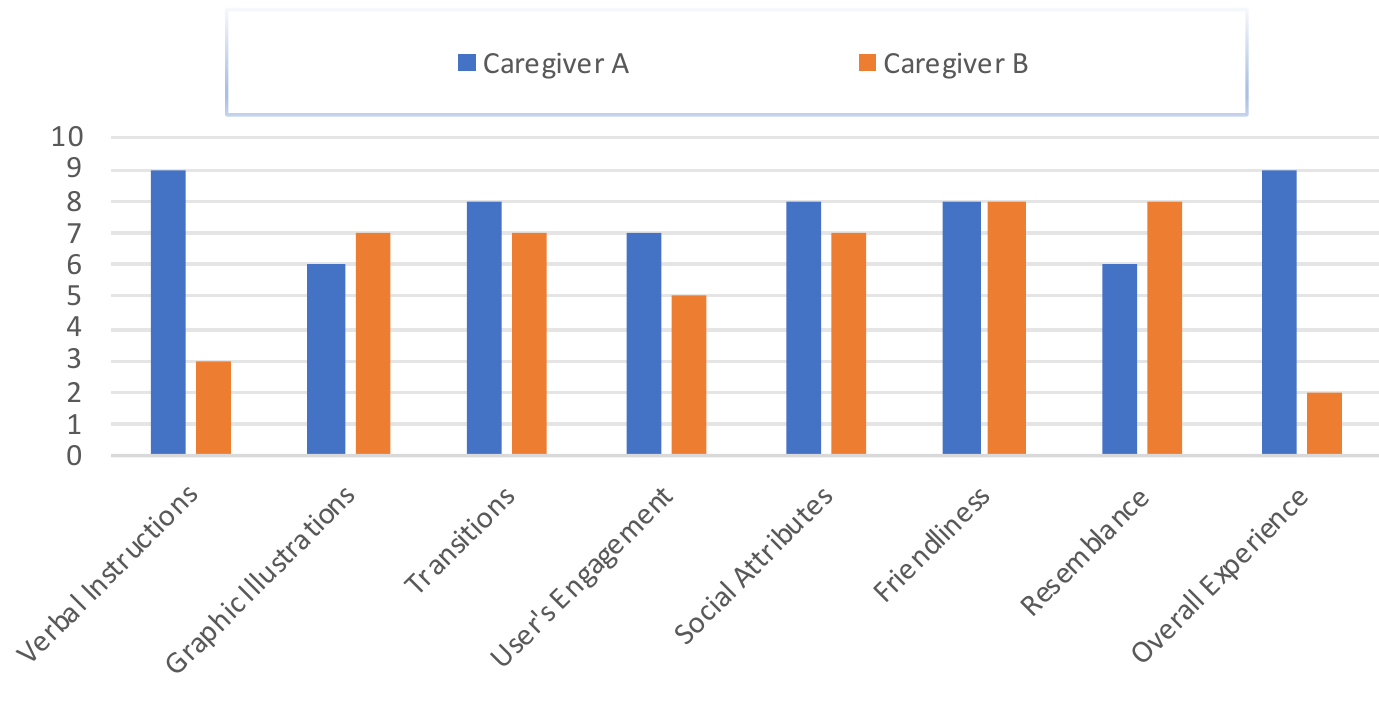}
\caption{Interview questionnaire feedback from caregivers}
\label{fig:results1}
\end{figure}

Both of the caregivers gave similar feedback (see Fig. \ref{fig:results1}) on most facets of the interaction design. It should be noted that there is a substantial divergence of opinion about \textit{Verbal Instructions} and \textit{Overall Experience} between these two caregivers. In the subsequent interview, one of them pointed out that the overall design neglected residents' incapabilities to reference. Specifically, an object or activity need to be available for residents right when the robot mentions it. Otherwise, it would pose a confusion for them due to their limited perception and understanding.

Overall, two caregivers' impressions of other aspects align with each other.

\subsection{User Study}

Table \ref{tbl:results2} summarizes the results of multiple matrix coding queries. This table has been categorized based on four dimensions: communication approach, type of information, form of interaction, and type of response. From the \textit{Interaction} section in Table 1, we note that the involvement of computer reduces the recurrence of disengagement. Based on our observation, residents tended to get distracted frequently by other residents or guests. They had got used to caregivers' company and thus they often ignored their caregivers. Unordinary objects, such as an iPad or computer, appeared to be more appealing to them.

Although the primary focus of this study is the visual aspects of robot design, we also notice the importance of verbal communication, which leads to positive feedback most of the time. Users were notably more responsive when a word or phrase corresponding to their interest is mentioned. This finding suggests that verbal and graphical stimuli could be combined together to maximize attraction. 

\begin{table}[hbt]
\centering
\caption{Matrix coding results of the user study}
\includegraphics[width=3.4in]{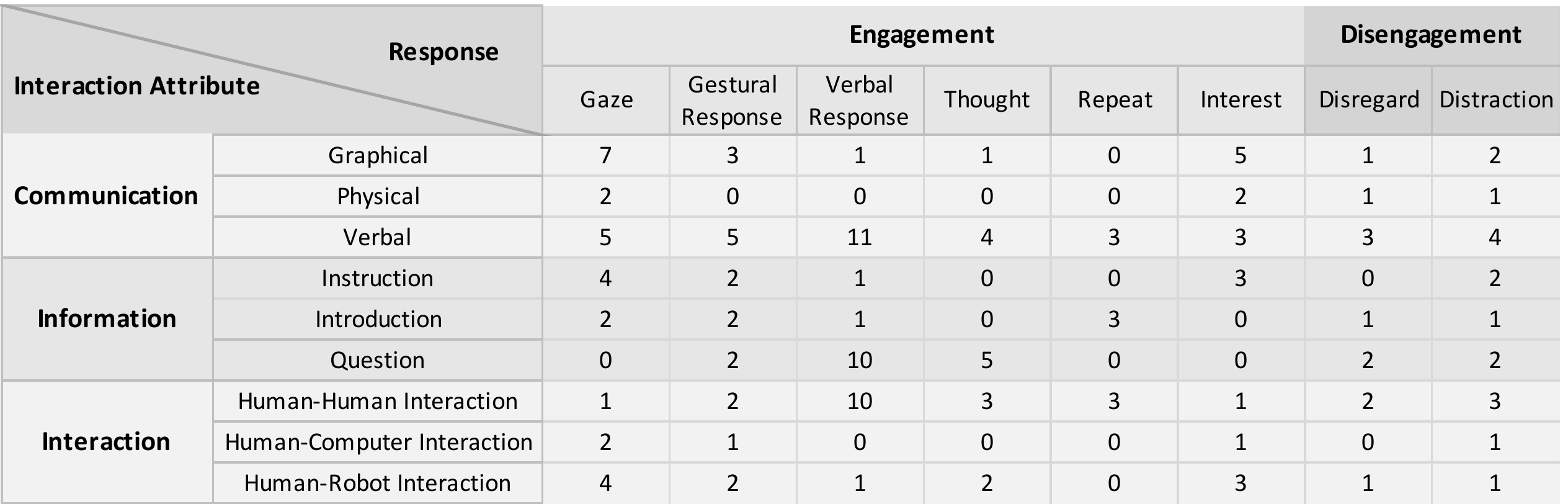}
\label{tbl:results2}
\end{table}

Amongst all types of information, \textit{question} seemed to be the most effective in terms of getting users' response. We noted an increasing chance of getting a response when the question was getting easier. For example, a user would respond to the questioner eight out of ten times when the question can be answered simply by a ``Yes" or ``No." When the robot was giving instructions, usually users would not ignore this information. However, oftentimes they got distracted by the objects that the robot was referring to. An example is that Resident 3 was occupied by the board game on the table while the robot was narrating the steps. She failed to grasp the visual information displayed on the screen. This occurrence indicates that their acceptance of step-by-step instructions is rather limited. Furthermore, users always processed the verbal information first, before perceiving others. This stands out as a significant pattern for designers to keep in mind. Different kinds of information may not be parallel all the time. Depending on user's perceptive abilities, there could be a layered and ordered relation among them. 

Gaze can be used as not only state-display cues on a robot to indicate the robot's conversational and operating states \cite{Breazeal2008}, but also a determining evidence to judge the extent of user's engagement. People in a social condition would gaze straight at the conversation partner, whereas in a less social condition they would turn away from the partner \cite{Heerink2009}. The GUI design embodied strong and concise visual hierarchies to attract and keep users' gaze most of the time. When users lost their focus on the screen, it might indicate an incompatibility or discontinuity of information from the previous scene. From the user study, we conclude that the GUI on the robot plays an essential role in guiding the users and in reinforcing the engagement.

\section{Discussion}

Our results indicate that the overall design of this primitive robot system was effective. We acknowledge that our design is not polished enough to handle certain scenarios at this stage. 

Through testing our prototype, we learned that many commonly accepted design principles could not be applied to our users. Animating transitions is an effective approach to reducing cognitive perturbation when a user is trying to follow a change in an adapting GUI \cite{Dessart2011}. Transitions we implemented in our prototype included resizing, fading, movement, and etc. These  transitions were designed to be smooth and gradual for users to cope with new contents, especially new types of contents. It is a widely accepted design principle that contrasty colours of subdued tone contribute substantially to the readability of textual and graphic information, and the gradual change of colours can help users shift their focus from one element or scene to another. However, this technique does not apply to individuals with severe cognitive impairments, as they could be struggling a lot with any subtle change of element. Hence, users would need extra time to process the change of colour before noticing the change of a pictorial element. 

The existence of the robot had also posed a challenge of dual reference to the residents. Dual reference refers to the capability of processing two information inputs simultaneously. Although ordinary users would usually not have issues with handling multiple information streams at the same time, it was the opposite way round to our users. Therefore, the robot's functions should be more intelligently integrated into the usage scenario in the future. In fact, \textit{Aether} was designed to able to store the location information of items, and this would make it possible to augment the virtual human-robot communication with objects from the reality.

The tasks need to be divided into baby steps, which means that each step should only refer to a single action. In the context of GUI design, there should be only primary action on each screen \cite{Interface}. The real challenge of applying this principle into our design is the short-time memory of our target users, who will get lost immediately if there is not a strong coherence and continuity of graphic information.

People with higher cognitive capabilities were more likely to be attracted by animated graphic elements, and they could still understand the visual information. Some residents with limited cognitive abilities, by contrast, might find the movements or transitions of elements disturbing and confusing. For them, reading a static image had already been a challenge. For this reason, we need to categorize users into two groups again, as we did for symbols.

\section{Conclusion}

The long-term objective of this research is to develop a socially assistive robot for people suffering from developmental disabilities. We proved some of our assumptions regarding interaction design in the context of human-robot interaction. 

By conducting a user study, we have sensed the enormous potential of social robots for improving the live quality of our users. We also have learned about the special needs and behaviour patterns of our users from their caregivers. The caregivers' insights are invaluable to our future designs as we reiterate the prototyping development.

\section{Limitations and Future Work}

A limitation in our study was the time delay between scenes. Although we made the process of presenting information fairly slow in our prototype, it was still not slow enough. The caregiver pointed out that on average an elderly with cognitive impairments need at least 50 seconds to process a piece of information. 

To enhance the replicability and to reduce the complexity, the WOZ experiments in the user study were semi-automatic, which meant that the wizard did not have maximum control of the prototype while the user study was ongoing. The wizard could only initiate or pause the prototype, and move the robot. There could not be any improvisation during the user study. This had been a bottleneck for the robot to adjust its pace to cater to users' sluggishness.

Based on this study, we plan to continue the iterative prototyping process, and design new GUI prototypes, including new scenarios like medication reminders. We will revise the user study procedures in order to obtain more accurate findings.

\section{Acknowledgement}
The authors thank JDQ Inc. for providing funding and assistance to this project, DDA for their support and arrangement for the user studies and interviews, and Professor Takeo Igarashi at the University of Tokyo for giving invaluable advice on this paper.


\begin{thebibliography}{10}

\bibitem{Bizier2015}
C.~Bizier, G.~Fawcett, and S.~Gilbert, ``{Canadian Survey on Disability , 2012
  Developmental disabilities among Canadians aged 15 years and older},''
  no.~89-654, p.~3, 2015.

\bibitem{DSOntario2016}
DSOntario, ``{What is A developmental disability? - developmental services
  Ontario},'' dec 2016.

\bibitem{Broekens2009}
J.~Broekens, M.~Heerink, and H.~Rosendal, ``{Assistive social robots in elderly
  care: a review},'' {\em Gerontechnology}, vol.~8, no.~2, pp.~94--103, 2009.

\bibitem{Baisch2017}
S.~Baisch, T.~Kolling, A.~Schall, S.~R{\"{u}}hl, S.~Selic, Z.~Kim, H.~Rossberg,
  B.~Klein, J.~Pantel, F.~Oswald, and M.~Knopf, ``{Acceptance of Social Robots
  by Elder People: Does Psychosocial Functioning Matter?},'' {\em International
  Journal of Social Robotics}, vol.~9, pp.~293--307, apr 2017.

\bibitem{Walters2008}
M.~L. Walters, D.~S. Syrdal, K.~Dautenhahn, R.~{Te Boekhorst}, and K.~L. Koay,
  ``{Avoiding the uncanny valley: Robot appearance, personality and consistency
  of behavior in an attention-seeking home scenario for a robot companion},''
  {\em Autonomous Robots}, vol.~24, no.~2, pp.~159--178, 2008.

\bibitem{Mohd2012}
W.~Zainon, W.~S. Yee, C.~S. Ling, and C.~K. Yee, ``{Exploring the Use of
  Cognitive Psychology Theory in Designing Effective},'' {\em International
  Journal of Engineering and Industries}, vol.~3, no.~September, pp.~66--74,
  2012.

\bibitem{Wang2017}
L.~Wang, W.~Chen, and H.~Li, ``{Use of 3D faces facilitates facial expression
  recognition in children},'' {\em Scientific Reports}, vol.~7, no.~April,
  pp.~1--6, 2017.

\bibitem{Feil-Seifer2007}
D.~Feil-Seifer, K.~Skinner, and M.~J. Matari{\'{c}}, ``{Benchmarks for
  evaluating socially assistive robotics},'' {\em Interaction Studies: Social
  Behaviour and Communication in Biological and Artificial Systems}, vol.~8,
  no.~3, pp.~423--439, 2007.

\bibitem{Kobuki2015}
Kobuki, ``{About | Kobuki},'' 2015.

\bibitem{Mubin2014}
O.~Mubin, T.~D'Arcy, G.~Murtaza, S.~Simoff, C.~Stanton, and C.~Stevens,
  ``{Active or passive?: Investigating the impact of robot role in meetings},''
  {\em IEEE RO-MAN 2014 - 23rd IEEE International Symposium on Robot and Human
  Interactive Communication: Human-Robot Co-Existence: Adaptive Interfaces and
  Systems for Daily Life, Therapy, Assistance and Socially Engaging
  Interactions}, pp.~580--585, 2014.

\bibitem{Hendrick1971}
C.~Hendrick and S.~R. Brown, ``{Introversion, extraversion, and interpersonal
  attraction.},'' {\em Journal of Personality and Social Psychology}, vol.~20,
  no.~1, pp.~31--36, 1971.

\bibitem{Brandon2012}
M.~Brandon, ``{Effect of Robot-User Personality on the Acceptance of Domestic
  Assistant Robots for Elderly},'' {\em Chemistry {\&} {\ldots}}, no.~June
  2011, pp.~1--104, 2012.

\bibitem{Leyzberg2011}
D.~Leyzberg, E.~Avrunin, J.~Liu, and B.~Scassellati, ``{Robots that express
  emotion elicit better human teaching},'' {\em Proceedings of the 6th
  international conference on Human-robot interaction - HRI '11}, p.~347, 2011.

\bibitem{Song:2017fj}
S.~Song and S.~Yamada, ``{Expressing Emotions through Color, Sound, and
  Vibration with an Appearance-Constrained Social Robot},'' in {\em the 2017
  ACM/IEEE International Conference}, (New York, New York, USA), pp.~2--11, ACM
  Press, 2017.

\bibitem{McTear2016}
M.~McTear, Z.~Callejas, and D.~Griol, {\em {Introducing the Conversational
  Interface}}.
\newblock 2016.

\bibitem{nvivo}
{QSR International}, ``{What is NVivo?},'' 2018.

\bibitem{Breazeal2008}
C.~Breazeal, A.~Takanishi, and T.~Kobayashi, ``{Social Robots that Interact
  with People},'' {\em Springer Handbook of Robotics}, pp.~1349--1369, 2008.

\bibitem{Heerink2009}
M.~Heerink, B.~Kr{\"{o}}se, B.~Wielinga, and V.~Evers, ``{Measuring the
  influence of social abilities on acceptance of an interface robot and a
  screen agent by elderly users},'' {\em BCS-HCI '09 Proceedings of the 23rd
  British HCI Group Annual Conference on People and Computers: Celebrating
  People and Technology}, no.~January, pp.~430--439, 2009.

\bibitem{Dessart2011}
C.-E. Dessart, V.~{Genaro Motti}, and J.~Vanderdonckt, ``{Showing user
  interface adaptivity by animated transitions},'' {\em Proceedings of the 3rd
  ACM SIGCHI symposium on Engineering interactive computing systems - EICS
  '11}, p.~95, 2011.

\bibitem{Interface}
J.~Porter, ``{Design Principles of User Interface}.''

\end{thebibliography}
\end{document}